%% file: paper.tex
\documentclass[letterpaper]{article} 
\usepackage{aaai2026}  
\usepackage{times}  
\usepackage{helvet}  
\usepackage{courier}  
\usepackage[hyphens]{url}  
\usepackage{graphicx} 
\usepackage{natbib}  
\usepackage{caption} 
\usepackage{algorithm}
\usepackage{algorithmic}
\usepackage{booktabs}
\usepackage{tabularx}
\usepackage{makecell}
\usepackage{listings}
\usepackage{newfloat}
\usepackage{xcolor}
\DeclareCaptionStyle{ruled}{labelfont=normalfont,labelsep=colon,strut=off} 
\floatstyle{ruled}
\newfloat{listing}{tb}{lst}{}
\floatname{listing}{Listing}
\title{
Simulating Human-Like Learning Dynamics with LLM-Empowered Agents}

\author{
    Yu Yuan\textsuperscript{\rm 1}\textsuperscript{\rm 2}\equalcontrib,
    Lili Zhao\textsuperscript{\rm 1}\textsuperscript{\rm 2}\equalcontrib,
    Wei Chen\textsuperscript{\rm 1}\textsuperscript{\rm 2} ,
    Guangting Zheng\textsuperscript{\rm 1}, \\
    Kai Zhang\textsuperscript{\rm 1}\textsuperscript{\rm 2} ,
    Mengdi Zhang\textsuperscript{\rm 3},
    Qi Liu\textsuperscript{\rm 1}\textsuperscript{\rm 2}\thanks{Corresponding author.},
}
\affiliations{
    \textsuperscript{\rm 1} University of Science and Technology of China \\
    \textsuperscript{\rm 2} State Key Laboratory of Cognitive Intelligence, China \\
    \textsuperscript{\rm 3} Meituan \\
    \texttt{\{yyhappier,liliz,chenweicw,zgt\}@mail.ustc.edu.cn} \\
    \texttt{\{kkzhang08,qiliuql\}@ustc.edu.cn} \\
    \texttt{zhangmengdi02@meituan.com}
}

\usepackage{bibentry}

\begin{document}

\maketitle

\begin{abstract}
Capturing human learning behavior based on deep learning methods has become a major research focus in both psychology and intelligent systems. Recent approaches rely on controlled experiments or rule-based models to explore cognitive processes. However, they struggle to capture learning dynamics, track progress over time, or provide explainability. To address these challenges, we introduce LearnerAgent, a novel multi-agent framework based on Large Language Models (LLMs) to simulate a realistic teaching environment. To explore human-like learning dynamics, we construct learners with psychologically grounded profiles-such as Deep, Surface, and Lazy-as well as a persona-free General Learner to inspect the base LLM's default behavior. Through weekly knowledge acquisition, monthly strategic choices, periodic tests, and peer interaction, we can track the dynamic learning progress of individual learners over a full-year journey. Our findings are fourfold: 1) Longitudinal analysis reveals that only Deep Learner achieves sustained cognitive growth. Our specially designed ``trap questions" effectively diagnose Surface Learner's shallow knowledge. 2) The behavioral and cognitive patterns of distinct learners align closely with their psychological profiles. 3) Learners' self-concept scores evolve realistically, with the General Learner developing surprisingly high self-efficacy despite its cognitive limitations. 4) Critically, the default profile of base LLM is a ``diligent but brittle Surface Learner"-an agent that mimics the behaviors of a good student but lacks true, generalizable understanding. Extensive simulation experiments demonstrate that LearnerAgent aligns well with real scenarios, yielding more insightful findings about LLMs' behavior.
\end{abstract}

\begin{figure}[t]
  \centering
  \includegraphics[width=0.96\linewidth]{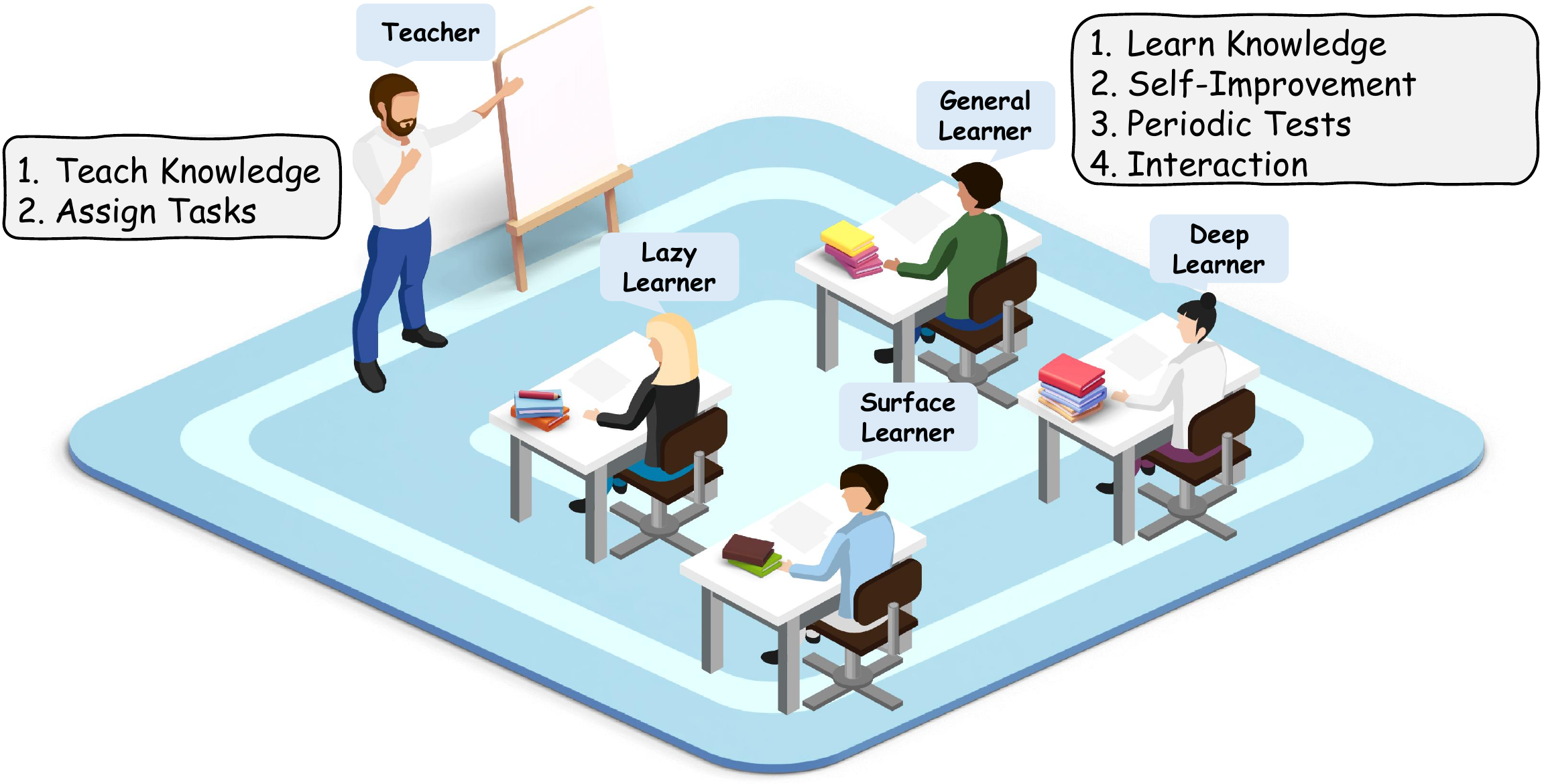}
  \caption{LearnerAgent conducts in-depth research on the learning processes of different learners through simulating real teaching scenarios.}
  \label{fig:simulation}
\end{figure}

\begin{figure*}[t]
  \centering
  \includegraphics[width=0.89\linewidth]{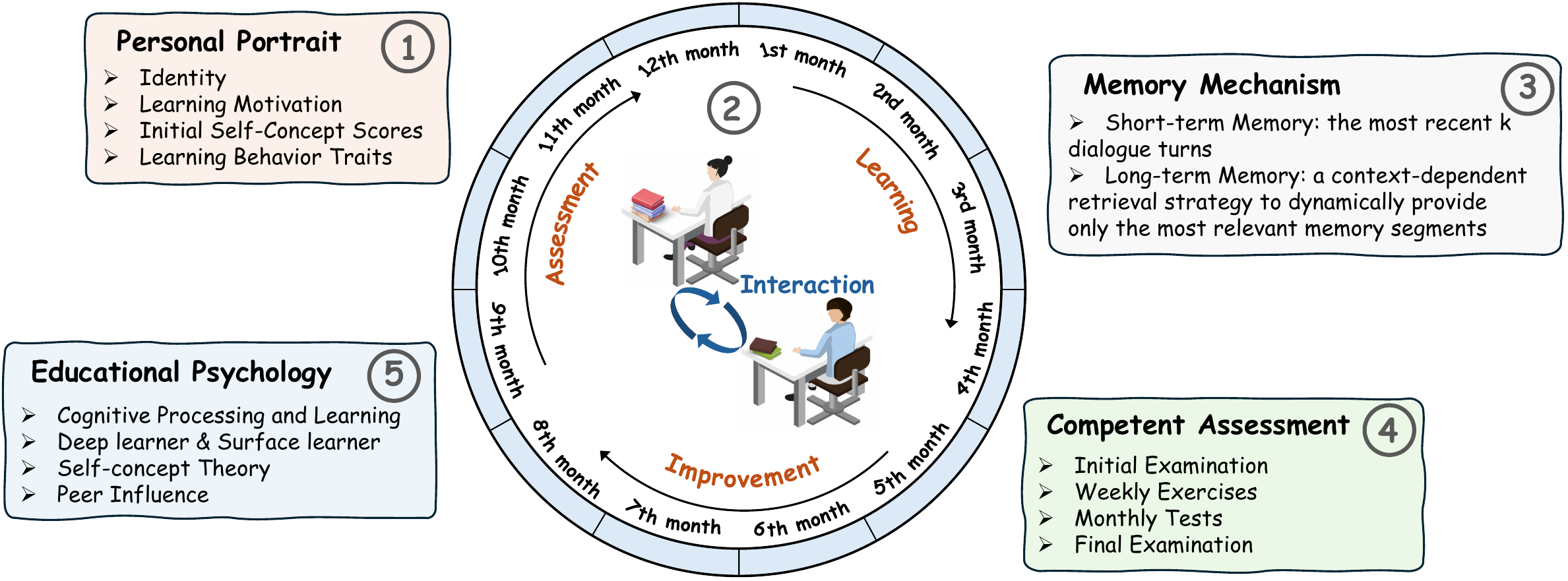}
  \caption{
  The LearnerAgent simulates a real teaching scenario by exploring how learners acquire knowledge and develop cognition through mechanisms such as personal profiling, adaptive learning, memory modeling, and performance assessment. Notably, some of its findings align with theories in Educational Psychology.
  }
  \label{fig:frame}
\end{figure*}

\input{sections/introduction}

\input{sections/related}

\input{sections/model}

\input{sections/experiment}

\input{sections/discussion}

\input{sections/conclusion}


\bibliography{aaai2026}

\newpage
\input{sections/appendix}

\end{document}

%% file: sections/introduction.tex
\section{Introduction}

As Large Language Models (LLMs) become increasingly prevalent in real-world applications \citep{weiemergent,guo2025deepseek,yuan2025cad}, researchers have begun to investigate the parallels and distinctions between model behavior and human learning patterns.
This stems from the recognition that studying these similarities and differences can help bridge the gap between artificial and human cognition \cite{park2023generative,tran2025multi}.
Specifically, Educational Psychology-with its robust theories on human learning processes \citep{marton1976qualitative, sweller2011cognitive}-offers valuable insights for deep learning systems. By applying these insights, researchers can explore a wide range of student learning behaviors (ranging from surface learning to deep learning patterns). It enables targeted optimization of the models' own learning mechanisms, ultimately improving their performance across critical downstream tasks, including Adaptive Learning \cite{liu2019exploiting}, and model capability diagnosis \cite{dong2025knowledge}.

Traditional cross-over research \citep{marton1976non, marsh1990self-concept,ezzaim2023ai} mainly relies on questionnaire surveys or rule-based measurement models to assess student learning performance, whether surface-level or deep-level. However, this approach has notable limitations: it is not only time-consuming and labor-intensive but also constrained by static inferences drawn from final responses from learners or model outputs. As a result, it struggles to provide dynamic and explainable evaluations of learning behaviors in complex and flexible environments.

Inspired by the success of LLM-empowered agents in simulating complex social systems—such as economic interactions \citep{zhao2023competeai}, natural science \cite{bran2023chemcrow}, and quantitative political analysis \citep{li2025political}, we propose a novel \textbf{LearnerAgent} framework to address the aforementioned challenges. As illustrated in Figure~\ref{fig:simulation}, LearnerAgent simulates real-world teaching environments. The teacher teaches knowledge and assigns tasks, while learners engage in learning and improvement over 12-month journey, complete exams, and interact with peers. 

To fully explore dynamic and diverse learning behaviors in real-world scenarios, we construct distinct profiles for learners from multiple dimensions as shown in Table \ref{tab:learner_profiles}.
Learners acquire knowledge weekly and make monthly strategic choices from knowledge consolidation and cognitive reflection. Periodic tests provide comprehensive assessment on their dynamic performance progress, while peer interaction and debate simulate social learning.

Through extensive experiments with LearnerAgent, we uncover several key findings:
\begin{itemize}
\item  LearnerAgent successfully simulates distinct learners (\textit{Deep}, \textit{Surface}, \textit{Lazy}) in a realistic teaching environment, with each showing behaviors aligned with their profiles in learning strategies, reasoning, and cognitive effort.
\item Longitudinal analysis shows that only the \textit{Deep} Learner achieves sustainable cognitive growth. The \textit{Surface} Learner's knowledge proves brittle, exhibiting shortcut learning behavior.
\item The self-concept of learners evolves dynamically. Profiled learners maintain stable self-perceptions, while the persona-free \textit{General} Learner develops an increasing self-efficacy, mirroring human-like confidence growth.
\item Learners respond differently to peer influence: \textit{Deep} Learner acts as a rational debater, \textit{Surface} Learner remains cognitively rigid, and \textit{Lazy} Learner is highly susceptible to persuasion.
\item The \textit{General} Learner (base LLM) defaults to a ``diligent but brittle surface learner'' style, mimicking ideal student behaviors while lacking deeper understanding, which reflects a reliance on shallow pattern-matching.
\end{itemize}

\begin{table}[t]
\small
\begin{center}
\setlength{\tabcolsep}{4.0pt} 
\renewcommand{\arraystretch}{1.05}
\begin{tabular}{llll}
\toprule
\bf Learner & \bf Motivation & \bf Self-Concept & \bf Dev. Strategies \\
\midrule
Deep & Intrinsic & High & Long-term \\
Surface & Extrinsic & Moderate & Test-oriented \\
Lazy & Minimal effort & Low & Passive \\
General & - & - & - \\
\bottomrule
\end{tabular}
\caption{Learners are defined along multiple dimensions, including Learning Motivation, Self-Concept, and Development (Dev.) Strategies, in addition to basic information.}
\label{tab:learner_profiles}
\end{center}
\end{table}

%% file: sections/related.tex
\section{Related Work}

\subsection{LLM-empowered Agents} 
Due to the powerful capabilities \cite{hurst2024gpt,comanici2025gemini} and human-like behaviors demonstrated by LLMs \cite{zhao2025evaluating}, several studies utilized the perception, planning, memory, and collaboration abilities of LLMs \cite{divband2022intelligent,fischer2021loop}
to simulate realistic scenarios and construct more intelligent agents.
Through coordination among different agents, Agentic AI can be classified into three types: Cooperation \cite{li2023theory,wang2024macrec}, Competition \cite{zhao2023competeai,liang2024encouraging,chen2023put}, and Coopetition \cite{abdelnabi2024cooperation,davidsonevaluating}.
Utilizing and exploring different types of coordination, agents have been widely used to simulate real-world implementations, including economic market \cite{zhao2023competeai,horton2023large,li2024econagent}, human behavior \cite{aher2023using,dillion2023can,park2023generative}, social science \cite{li2025political,argyle2023out,park2022social,gurcan2024llm,qian2023communicative}, and natural science \cite{bran2023chemcrow,boiko2023autonomous}.

Drawing on key findings from Educational Psychology \citep{chin2000learning, ryan2000motivation}, such as the progression from novice to expert—where surface-level strategies decrease while deep-level strategies increase (\textbf{cognitive development}) \cite{alexander2004model}; the role of \textbf{self-concept} as a central driver of motivation and behavior~\cite{marsh1990self-concept}; and the effect of \textbf{peer influence} on academic performance~\citep{delay2016peer}, we simulate a realistic teaching environment. This allows us to explore human cognitive development and learning behavior in greater depth, offering a novel perspective on how these psychological factors manifest in interactive settings.

\subsection{Shortcut Learning behavior of LLMs}
Due to the influence of training methods, data, and model architectures, LLMs, during the training process, rely on shortcuts to learn the spurious correlations between certain features, resulting in hallucinations and errors in the inference process \cite{weng2024reward}.
This behavior is similar to the shortcut learning exhibited by humans, where they quickly find shortcuts to solve current problems without delving into the underlying issues \cite{NMI20,zhao2024comi,zhao2025mimu}.
Several studies have revealed the shortcut learning behaviors in LLMs. \citet{yuan2024llms} found LLMs relied on different shortcuts such as Lexical Overlap, Subsequence, Constituent, Negation, Position, and Style, leading to hallucinations and failure in downstream tasks. \citet{lazyllm-acl23} pointed out that LLMs act as lazy learners, giving up deep thinking by directly learning the potential spurious correlations within the context examples.

In this paper, we draw a parallel between human and model learning. By simulating learners through a multi-agent system, we focus on their learning development, i.e., deep learning, shortcut learning, and lazy learning.

%% file: sections/model.tex
\section{LearnerAgent Framework}

LearnerAgent constructs a realistic learning environment through role-playing simulation as shown in Figure \ref{fig:frame}, which simulates the full-year learning journey based on theories of Educational Psychology and Cognitive Science \citep{ryan2000motivation, marsh2011self-concept, delay2016peer}.
It consists of two primary roles: a Teacher Agent and multiple Learner Agents, each with a distinct profile construction.

\subsection{Profile Construction}
\noindent\textbf{Teacher Agent ($T$)}, as an experienced high school instructor, is responsible for delivering knowledge and assigning tasks, including organizing teaching activities, evaluating learner performance, and providing appropriate guidance.

\noindent\textbf{Learner Agents} ($L = \{L_d, L_s, L_l, L_g \}$) are categorized into four predefined types, with their distinct behavioral patterns derived from individual role profiles. In the full-year (12-month) learning, all learners primarily focus on knowledge acquisition and periodic assessments.

To simulate different learners that align with real-life scenarios, learners are depicted as various profiles based on three dimensions in addition to basic information, as  summarized in Table~\ref{tab:learner_profiles}: 
1) \textit{Learning Motivation}: refers to the drive to learn, including intrinsic curiosity, external rewards (e.g., grades), and a desire to avoid efforts;
2) \textit{Initial Self-Concept Score}: represents learners' assessment of their own abilities, which is reflected not only in their confidence levels but also in how much they are affected by evaluations (such as examination scores) and by their peers;
3) \textit{Development Strategies}: denotes the methodology learners adopt in different learning processes, aligning with their learning motivation and self-concept.

Based on these dimensions, we classify three distinct learners: Deep Learner ($L_d$), Surface Learner ($L_s$), and Lazy Learner ($L_l$). Moreover, General Learner ($L_g$), initialized only with Basic Information, serves as an essential baseline for comparison. 
Notably, $L_g$ can serve as a probe for observing the emergent behavioral tendencies of the underlying LLM itself. 
The detailed configurations for these learners can be found in Appendix A.

\subsection{Learning and Improvement}
\label{learning-and-improvement}
To explore the cognitive development process of learners, a structured cycle of learning, improvement, and assessment is well-designed through a one-year learning period.
Specifically, the initial baseline exam and final exam at the end of cycle are designed for assessing overall knowledge acquisition and growth. 
In each month, learners engage in learning and practice during the first three weeks, and dedicate the fourth week to summary, review, and take a monthly exam.

\noindent\textbf{Weekly Learning and Strategic Choices.} 
In the first three weeks of each month, learners follow a structured routine: studying provided materials, taking notes, and completing weekly tests. After each test, the $T$ provides standard explanations for all questions. 
At the end of three weeks, learners need to make two key choices that reflect self-regulation:
1) \textit{Knowledge Consolidation}: Summarize their learning notes from all three weeks into a single document to prepare for the monthly exam, or take a break.
2) \textit{Cognitive Reflection}: Reflect on their performance across the three weekly tests, identifying key insights and mistakes, or take a break.

\noindent\textbf{Monthly Review and Assessment.}
The fourth week of each month focuses on final review and assessment. Before the monthly exam, learners should make a final choice: to review their accumulated materials (e.g., summaries, reflections) thoroughly, or to rest.

\noindent\textbf{Peer Interaction and Debate.}
To simulate social learning and peer influence, LearnerAgent introduces a dynamic debate mechanism after each monthly exam. When two learners, $L_i$ and $L_j$, provide different answers to the same question $q$, they are prompted to engage in a debate. During the discussion, learners can defend their reasoning, challenge peers, or revise their initial answers. The teacher organizes the debate and determines its structure and number of rounds. The debate ends when any of the following conditions are met:
1) A learner explicitly changes their answer after being persuaded;
2) Both learners maintain their own views with well-supported reasoning;
3) Their opinions converge or become repetitive;
4) The number of debate rounds has become relatively high (exceeds $k_d$ rounds).

\subsection{Memory Mechanism}
Given that learning is inherently sequential and that LearnerAgent provides diverse scenarios to foster ongoing learner improvement, we employ both short-term and long-term memory mechanisms to enable effective knowledge acquisition, assessment, and reflection.

\noindent\textbf{Short-Term Memory} maintains immediate conversational context by storing the most recent $k$ dialogue turns in a fixed-size queue. This enables coherent and context-aware interactions with the teacher and peers.

\noindent\textbf{Long-Term Memory} serves as a persistent, structured store of the learner’s complete learning and improvement history. It archives key memories—including knowledge summaries, exam answers, self-reflections, debate records, monthly exam scores, and self-concept scores—as distinct entries for long-term retrieval.
To support personalized assistance and temporal alignment, we introduce a \textbf{context-dependent retrieval strategy} in which the framework dynamically provides the agent with only the most relevant memory segments, based on the current time step or learning phase.
As shown in Figure~\ref{fig:frame}, the strategy adapts to key learning stages by retrieving tailored memories:

1) Weekly learning retrieves topical summaries.
2) Monthly exams provide prior knowledge summaries, reflections, relevant past exam answers, and teacher feedback.
3) Debates include similar past responses and discussion threads.
4) Self-concept evaluation recalls prior self-concept scores and performance comparisons.
5) Final exams offer a consolidation of knowledge across the entire year.

\subsection{Competent Assessment}
To fully evaluate learners’ knowledge and ability development throughout the learning process, we employ a comprehensive assessment strategy covering both academic performance and psychological changes.

\noindent\textbf{Performance Evaluation.} In addition to the initial, weekly, monthly, and final exams in the learning cycle, the \textbf{\textit{Trap Questions}} are constructed to evaluate the deeper understanding of learners beyond rote learning. 
Similar to \citep{yuan2024llms}, we design these trap questions with the same structure as weekly practice questions but different reasoning due to subtle context changes, to deeply diagnose and uncover learners’ behavior patterns.
We use a powerful closed-source LLM to generate them, and then manually annotated and filtered as shown in Figure~\ref{fig:shortcut}.

\noindent\textbf{Psychological Evaluation.} 
To inform better self-evaluation of different learners, they are provided with two key inputs: 1) their historical self-concept scores, and 2) exam scores of other learners. They are then asked to update their self-concept scores (e.g., on a 0–100 scale) by considering their learning capability, knowledge mastery, progress, and their perception of their position relative to classmates.

%% file: sections/experiment.tex
\section{Experimental Setups}
In this section, we conduct extensive experiments to evaluate the effectiveness of LearnerAgent and answer the following Research Questions (RQs).
\begin{itemize}
    \item \textbf{RQ1}: What knowledge acquisition and cognitive development processes do learners exhibit during learning journey? Can they solve the trap questions?

    \item \textbf{RQ2}: What distinct behavioral and cognitive patterns are associated with each learner?
     
    \item \textbf{RQ3}: 
    How do learners’ perceptions of themselves (self-concept) evolve throughout learning processes?

    \item \textbf{RQ4}: How do learners respond to peer influence in interactive debates?
  
    \item \textbf{RQ5}: What is the emergent learning profile of a base LLM—a learner without a predefined persona (as the General Learner)?

\end{itemize}

\begin{figure*}[t]
  \centering
  \includegraphics[width=\linewidth]{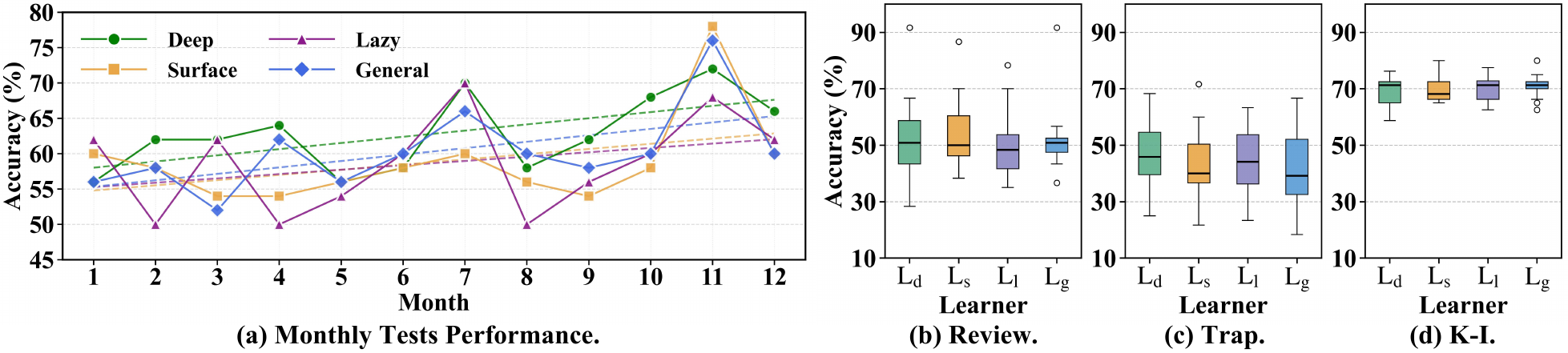}
    \caption{
    Development of learner performance. 
    \textbf{(a)}: Total scores of the four learners over 12 months, covering all question types. The dashed lines represent the linear trend for each learner's scores;
    \textbf{(b)}: Accuracy on \textit{review} questions, targeting short-term retention of current content;
    \textbf{(c)}: Accuracy on \textit{trap} questions, which assess transfer learning and deeper understanding;
    \textbf{(d)}: Accuracy on \textit{knowledge-integration (K-I)} questions, evaluating cumulative knowledge over time.
    }
  \label{fig:development}
\end{figure*}
\begin{figure}[t]
  \centering
  \includegraphics[width=0.95\linewidth]{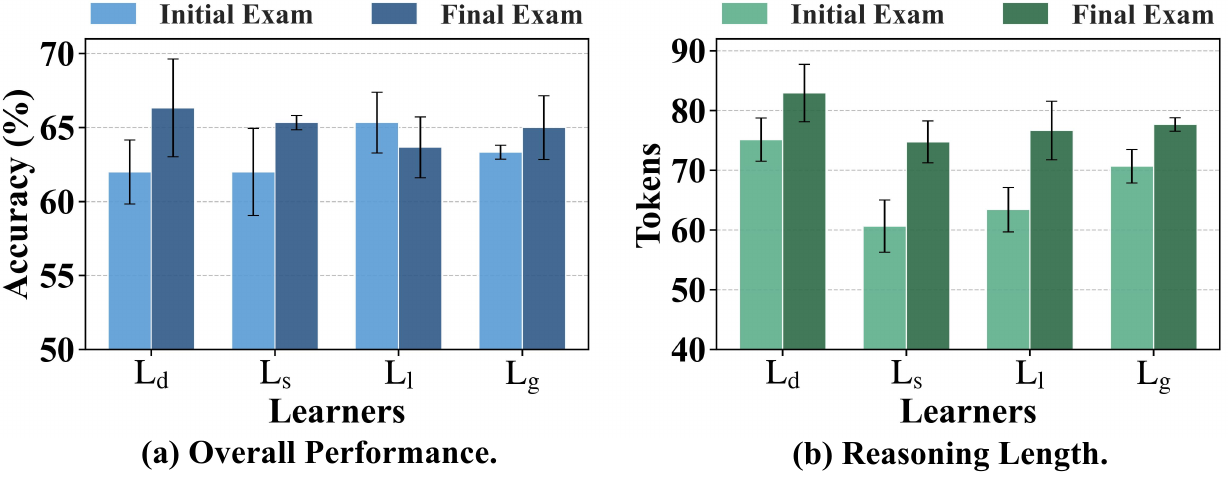}
  \caption{The comparison of overall performance \textbf{(a)} and reasoning abilities \textbf{(b)} across the initial to the final exams.}
  \label{fig:comparison}
\end{figure}

\subsection{Test Suite Construction}

We formulate the learning task as English grammar acquisition, a natural testbed for evaluating knowledge mastery, deep reasoning of learners. 
To support this, we manually construct a new test suite to simulate a year-long grammar learning process.
All materials are collected from authoritative Gaokao\footnote{The Gaokao is China’s National College Entrance Examination. Its English section emphasizes systematic grammar skills.} English preparation resources, including high school textbooks, practice books, and official compilations of past exams, ensuring both authenticity and comprehensive grammar coverage\footnote{The materials are collected from gaokao.eol.cn}.

The test suite consist of two key components:
1) \textbf{Knowledge Points}: Each week introduces specific grammar concepts (e.g., verb tenses, clause types, sentence structure), aligned with the sequencing in Gaokao curricula. 
2) \textbf{Test Questions}: For each concept, we provide diverse question formats—multiple choice, fill-in-the-blank, and error correction—all designed to assess learners' ability to apply grammar rules flexibly. 
To further evaluate deep understanding beyond rote memorization, we construct \textit{Trap Questions}: automatically generated using Gemini-2.5-Pro~\citep{comanici2025gemini} and manually verified. These questions are structurally or lexically similar to past-learned questions, but introduce subtle contextual shifts, requiring learners to comprehend their knowledge in depth.
Test questions are introduced in the following periodic tests:
\begin{itemize}
    \item \textbf{Initial and Final Exams}: Same set of 100 questions designed to measure overall grammar proficiency at the beginning and end of the learning period.
    \item \textbf{Weekly Exercises}: 20 questions each week to review and consolidate newly acquired grammar topics.
    \item \textbf{Monthly Tests}:  A total of 12 exams, each comprising 50 questions distributed across three sections: (1) 15 \textit{Review} questions targeting that month’s topics to evaluate short-term retention; (2) 15 \textit{Trap} questions to assess deep understanding and transfer learning; and (3) 20 \textit{Knowledge-Integration (K-I)} questions integrating all covered knowledge points to measure cumulative proficiency.
\end{itemize}

\subsection{Implementation Details}
We employ Qwen-2.5-72B-Instruct \citep{qwen2} as the Teacher Agent to provide instruction and feedback, and Qwen-2.5-7B-Instruct for all Learner Agents. Each learner (Deep, Surface, Lazy, General) is configured with distinct profiles, and each configuration is evaluated over 3 independent runs to ensure robustness. Experiments are conducted on four NVIDIA A800-80GB GPUs. Detailed prompt settings are available in Appendix A, and more experimental results and analysis can be found in Appendix B.

\section{Experimental Results and Analysis}
\label{sec-analysis}

We conduct a series of detailed analyses of the learning and development processes of different learners in various scenarios based on the four research questions presented.

\subsection{Cognition and Learning Development (RQ1)}
We track the longitudinal development of each learner by analyzing three aspects: 1) their monthly exam performance over 12 months, 2) their performance on diverse question types (e.g., review, trap, and knowledge-integration (K-I)) to assess cognitive gains, and 3) their overall performance and reasoning ability from the initial to the final exams.

\noindent\textbf{Longitudinal Performance.} 
As shown in Figure~\ref{fig:development} (a), all learners show a clear upward trend in monthly test scores across the 12-month learning, though with considerable volatility.
Overall, the Deep Learner and General Learner show a somewhat different pattern of improvement, while the Lazy Learner and Surface Learner perform worst in most months. Among these, Lazy Learner shows the greatest fluctuation in performance.

\noindent\textbf{Cognition Across Question Categories.}
To reveal fine-grained cognitive strengths and weaknesses, we evaluate how distinct learners perform on diverse questions, as shown in Figure \ref{fig:development} (b–d).

On \textbf{\textit{Review}} questions assessing current-month knowledge, Surface Learner performs the best, reflecting a strategy well-suited to short-term memory and immediate assessments. Deep and General Learners achieve moderate performance, while Lazy Learners perform the weakest, consistent with their limited engagement with recent learning.

\begin{figure}[t]
\centering
\includegraphics[width=0.95\linewidth]{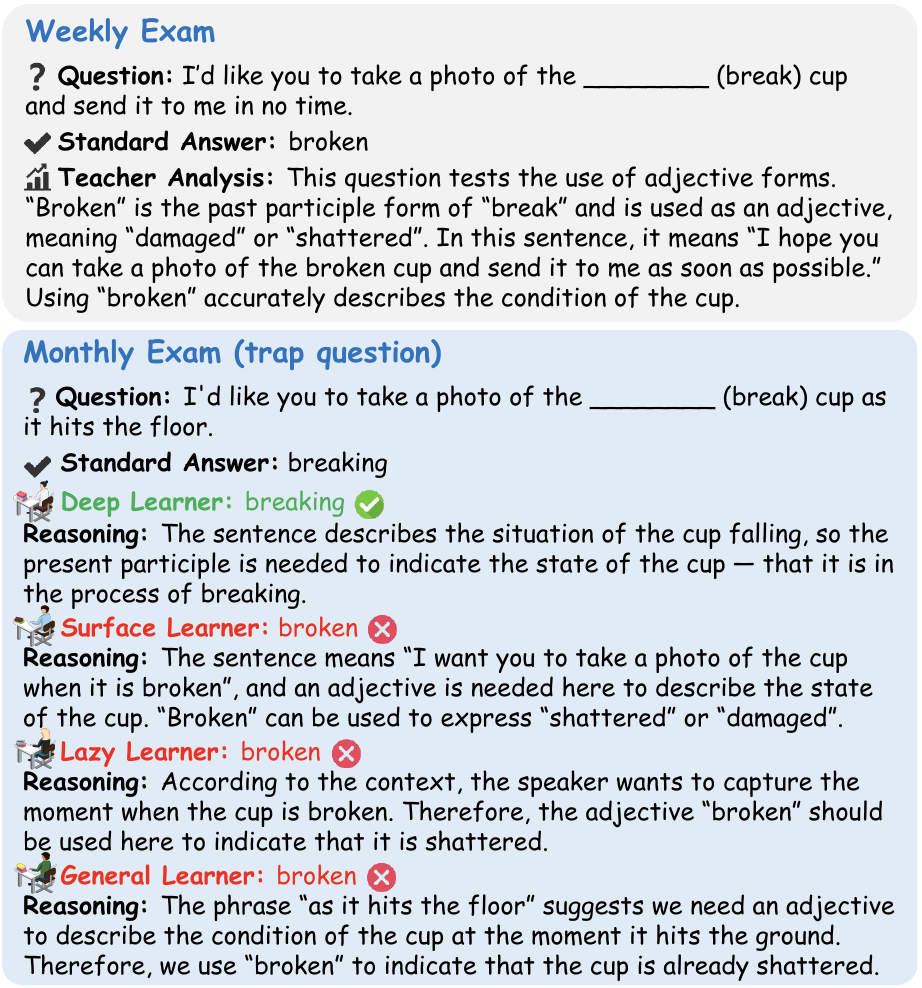}
\caption{A case study of a trap question distinguishing deep understanding from rote learning. By introducing a subtle contextual change to a familiar weekly item, this case illustrates that only the Deep Learner successfully adapts reasoning, while others rely on superficial pattern matching.}
\label{fig:shortcut}
\end{figure}

On \textbf{\textit{Trap}} questions designed to assess deep understanding and knowledge mastery, Deep Learner achieves the highest accuracy, demonstrating a strong ability to generalize principles rather than rely on rote learning. In contrast, Surface and General Learners struggle the most, reflecting their limited capacity for deep and true reasoning. 
Despite performing well on monthly review questions, their poor performance on trap questions reveals superficial and fragile knowledge mastery. As shown in Figure~\ref{fig:shortcut}, a trap question where the correct answer changed from ``broken" in the previous question to ``breaking", only Deep Learner answers correctly, while the Surface, Lazy, and General Learners reuse the incorrect prior answer. Deep Learner demonstrates contextual and grammatical reasoning, whereas Surface and Lazy Learners rely on rote memory, and General Learner shows partial adaptability but default to surface-level cues.

On \textbf{\textit{Knowledge-Integration}} questions testing cumulative knowledge, all learners perform similarly well. General Learner achieves the highest average score, making it the best knowledge accumulator.

\noindent\textbf{Overall Performance and Reasoning.} Through the direct comparison between initial and final exams as shown in Figure~\ref{fig:comparison}, the performance of all other learners improves except for Lazy Learner. All learners increase their reasoning length over time, suggesting a tendency toward more elaborate explanations. Deep Learner consistently shows the longest reasoning and steady gains in both length and performance. In contrast, Surface and Lazy Learners, despite large increases in reasoning length, remain the least elaborate reasoning overall, reflecting shallow and superficial responses. General Learner improves moderately on both metrics. LearnerAgent can effectively simulate diverse learning journey, and true cognitive growth depends on maintaining depth and quality—not just longer responses.

\begin{figure}[t]
  \centering
  \includegraphics[width=\linewidth]{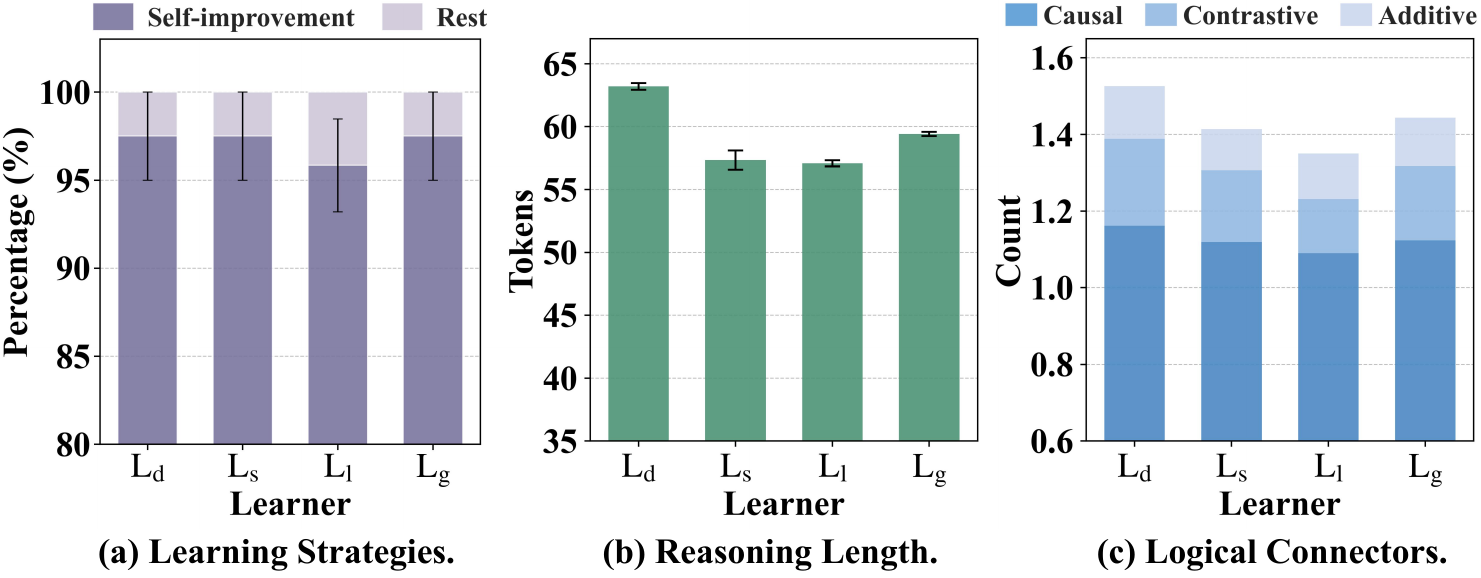}
  \caption{Behavioral and cognitive differentiation among learners. 
\textbf{(a)} Learning Strategies: Proportion of time allocated to self-improvement or rest. 
\textbf{(b)} Reasoning Length: Average number of tokens per reasoning. 
\textbf{(c)} Logical Connectors: Average number of logical connectors per reasoning and their distribution by type (causal, contrastive, additive). }
  \label{fig:behaviour}
\end{figure}

\subsection{Behavioral and Cognitive Patterns (RQ2)}




We make detailed analyses on distinct behavioral and cognitive patterns in each learner, based on their learning strategies and reasoning quality throughout the learning process.

\noindent \textbf{Learning Strategies.}
We analyze strategic choices of different learners between self-improvement (e.g., summarizing, reflecting, or reviewing) and rest when facing monthly exams described in Section~\ref{learning-and-improvement} by calculating their study-to-rest ratios.
As shown in Figure~\ref{fig:behaviour} (a), all learners mostly choose to study, reflecting an overall tendency for active engagement rather than rest. Lazy Learners, however, rest slightly more, suggesting lower motivation and efforts.

\noindent \textbf{Reasoning Patterns.}

To further analyze cognitive behaviors across learners, we conduct experiments on reasoning responses along two key dimensions: (1) \textit{Reasoning Length} (measured by average number of tokens per explanation), and (2) \textit{Logical Connector Density} (the frequency and variety of logical connectors, e.g., cause-effect, contrast, addition, used to structure reasoning).

First, reasoning length reveals clear differences in elaboration depth, as shown in Figure~\ref{fig:behaviour} (b). Deep Learner consistently produces the longest explanations, suggesting strong cognitive engagement and a focus on explanation. Surface and Lazy Learners, in contrast, produce significantly shorter responses, reflecting minimal effort or surface-level recall.

Second, the use of logical connectors reveals significant differences in reasoning complexity, as shown in Figure~\ref{fig:behaviour} (c). The Deep Learner has the highest connector density and the Lazy Learner the lowest. Deep Learner employs more contrastive connectors, a strong indicator of higher-order thinking that reflects the ability to weigh alternatives and reason nuances. In contrast, other learners use fewer contrastive connectors and rely more on simple causal connectors, indicating a more linear thinking style.

\subsection{Self-Concept Evolution (RQ3)}

\begin{figure}[t]
  \centering
  \includegraphics[width=0.96\linewidth]{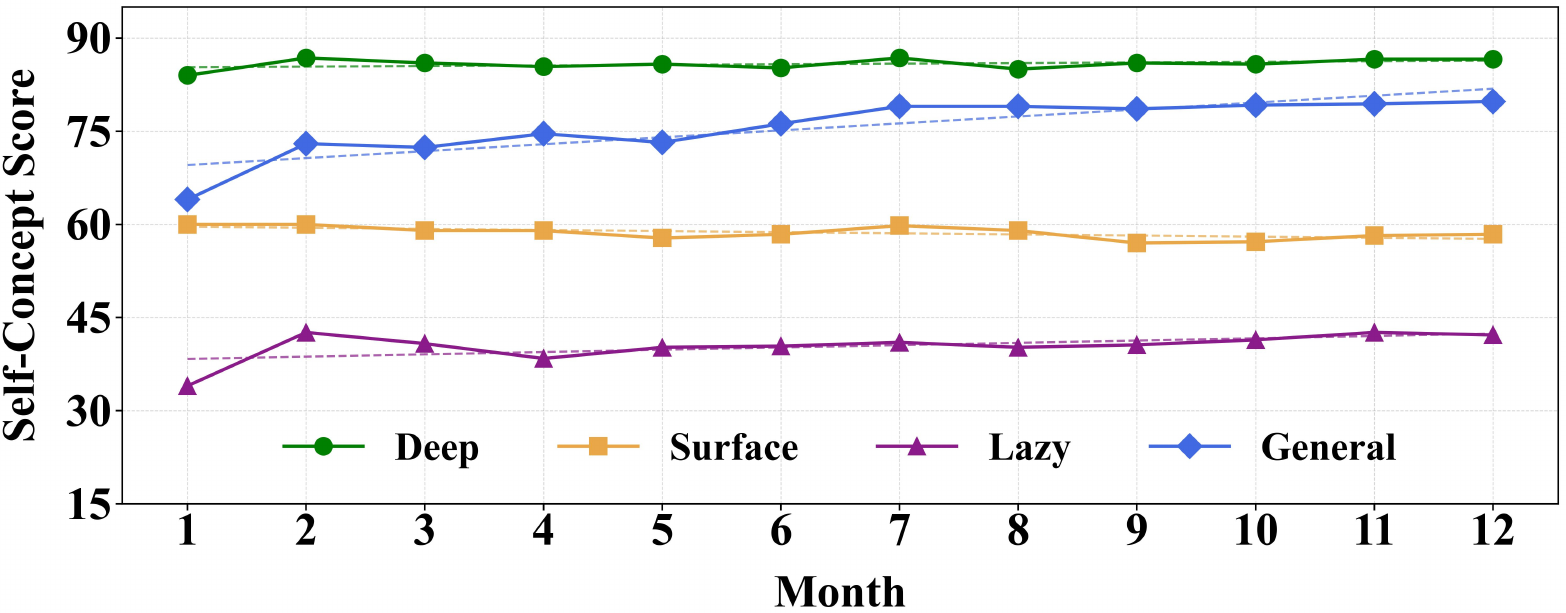}
  \caption{The evolution of self-concept scores for learners throughout 12-month learning.}
  \label{fig:self-concept}
\end{figure}

We track how each learner’s self-concept evolves over a 12-month period to examine how their perception changes in response to long-term learning and performance feedback. As shown in Figure~\ref{fig:self-concept}, learners with predefined personas generally mostly keep self-concepts in line with their initial profiles, though some subtle changes occur.

Deep Learner maintains consistently high and slightly rising self-concept scores, reflecting stable confidence largely unaffected by short-term performance changes or peer comparison. Surface Learner shows a slight decline, suggesting a more fragile self-view influenced by variable outcomes. Lazy Learner, who starts with the lowest self-concept, shows modest improvement, likely from small cumulative gains.
Each learner follows a distinct developmental trajectory, demonstrating LearnerAgent’s ability to simulate realistic learning journeys.
Notably, General Learner shows a special pattern: it begins without a predefined profile, starts at a moderate self-concept level, and exhibits a strong, steady increase over time. 
This emerging trend reflects how the base LLM gradually forms a dynamic self-concept that is at times overly optimistic.

\subsection{Peer Influence (RQ4)}

To evaluate how learners respond to peer influence during discussions, we define three metrics to capture interaction dynamics: 1) \textit{Persuasion}, the proportion of debates where the learner successfully persuades their peer, 2) \textit{Resist Wrong}, the rate at which the learner correctly rejects incorrect peer arguments, and (3) \textit{Accept Correct}, the rate at which the learner revises incorrect beliefs when given correct input. The results are summarized in Table~\ref{debate}.

Deep Learner performs best in both Persuasion and Resist Wrong rate, while also maintaining a high Accept Correct score. This combination reflects a learner that is both persuasive and resilient—capable of defending its views against misinformation while staying open to valid corrections. Surface Learner shows strong resistance to incorrect arguments and effectively accepts correct ones, yet it is the least persuasive, with a Persuasion rate as low as 7.13\%. Lazy Learner performs weakest overall, with the lowest scores in both Persuasion and Resist Wrong rate, making it especially prone to misinformation. Although its Accept Correct rate is moderate, its engagement remains shallow and inconsistent.
General Learner is surprisingly effective at arguing its views, achieving the second-highest Persuasion Rate of 12.0\%. However, its relatively low Resist Wrong score and Accept Correct rate indicate a weaker ability to resist incorrect arguments and accept valid feedback, leaving it vulnerable to misinformation from peers.

\begin{table}[t]
\small
\begin{center}
\setlength{\tabcolsep}{2pt} 
\renewcommand{\arraystretch}{1.05}
\begin{tabular}{lccc}
\toprule
\textbf{Learner} & \textbf{Persuasion  $\uparrow$} & \textbf{Resist Wrong $\uparrow$} & \textbf{Accept Correct $\uparrow$} \\
\midrule
Deep ($L_d$)    & \textbf{15.0} & \textbf{98.4}  & 19.2 \\
Surface ($L_s$) & 8.38 & 96.4 & \textbf{22.1} \\
Lazy ($L_l$)    & 7.13 & 86.9 & 17.1 \\
General ($L_g$)  & 12.0 & 90.9 & 19.1 \\
\bottomrule
\end{tabular}
\caption{
Debate performance is assessed over 12 months using: Persuasion (successfully persuading peers), Resist Wrong (resisting incorrect peer input), and Accept Correct (updating beliefs in response to correct arguments).
}
\label{debate}
\end{center}
\end{table}

\subsection{Emergent Profile of the General Learner (RQ5)}
As a learner without a pre-defined profile, it is important to investigate the General Learner's behavior across all tasks, which can be regarded as the learning trajectory of the base model.
Specifically, at first glance, the General Learner appears highly engaged: it frequently chooses self-improvement activities, provides reasonably detailed answers, and excels at knowledge-integration questions, with performance second only to the Deep Learner. However, a closer look reveals a disconnect between its outward diligence and actual learning depth. Although its self-concept rises steadily to match the Deep Learner’s high confidence, this confidence is not supported by performance on challenging tasks—especially Trap Questions, where its accuracy is nearly as low as the Surface Learner’s. Its weaker debate performance, with less persuasive arguments and limited reasoning, further reflects this gap. 
The results indicate that the General Learner adopts a profile best described as a ``\textbf{diligent but brittle surface learner}''.

In conclusion, the base LLM does not emerge as a capable deep reasoner. Instead, it relies on surface-level strategies, adopting the habits of a diligent but overly confident learner while struggling with deeper, more flexible reasoning. These findings highlight the need for cognitive guidance—such as a structured environment and explicit instructional support—to help LLMs move beyond pattern-matching and develop more robust, human-like deep understanding.

%% file: sections/discussion.tex
\section{Discussion}

\begin{table}[t]
\small
\begin{center}
\setlength{\tabcolsep}{2pt} 
\renewcommand{\arraystretch}{1.2}
\begin{tabular}{cccc}
\toprule
\textbf{System} & \textbf{Goal} & \textbf{Self-evaluation} & \textbf{Learning Strategies} \\
\midrule
Learner & \makecell{Test-oriented} & Self-concept & Surface  Learning \\
D-L    & Minimize Loss & Confidence & Fine-Tuning \\
LLMs    &  Minimize Loss  & Confidence & \makecell{Reinforcement \\Learning} \\
\bottomrule
\end{tabular}
\caption{
Comparison of learning behaviors across surface learners, Deep Learning (D-L) models, and LLMs.
}
\label{tab:learning_comparison}
\end{center}
\end{table}

As presented in Table~\ref{tab:learning_comparison}, we compare learners, deep learning models, and LLMs, and find that despite their \textit{differing learning goals, evaluation methods, and strategies, all are prone to shortcut learning}. Whether driven by test performance or loss minimization, their short-term-focused strategies often lead to behaviors such as exploiting spurious feature-label correlations during fine-tuning or aligning with human biases in reinforcement learning. Moreover, their evaluation methods often reflect overconfidence. In this paper, we uncover the behavior exhibited by human-like learners and LLMs, particularly the General Learner—a persona-free base LLM that employs shortcut learning (e.g., diligence and detailed answers) but fails on tasks requiring deep, generalizable understanding.

Furthermore, LearnerAgent provides a novel simulation framework with broad implications for AI and Education domains. For AI researchers, it offers a psychologically grounded approach to identify and mitigate shortcut learning. For Education scholars, it serves as a dynamic alternative to static surveys, enabling the study of learning and development over time. Future research can build on this framework to promote deep learning, mitigate the gap between self-concept and achievement, and benchmark emergent cognitive tendencies across diverse LLMs.

%% file: sections/conclusion.tex
\section{Conclusion}
In this paper, we introduced \textbf{LearnerAgent}, a multi-agent framework designed to simulate and analyze the complex dynamics of human learning based on Educational Psychology. 
Through a year-long simulation, we successfully tracked diverse learning behaviors, developmental trajectories, and social interactions across distinct learners.
Our work yielded three key insights. First, persona-driven agents could effectively replicate nuanced human learning behavior with high fidelity. Second, we identified distinct learning deficits among agents that exhibited similar short-term performance but diverged significantly in long-term generalization—a finding illustrated by performance ``traps'' that expose deeper vulnerabilities.
Third, the default emergent behavior of a persona-free base LLM, termed the ``diligent but brittle surface learner,'' reflected an agent that mastered surface competence yet lacked robust, generalizable understanding. We further showed that an inflated self-concept contributed to overconfidence, ultimately hindering growth.

%% file: sections/appendix.tex
\appendix
\section{Appendix A: Detailed Implementations}
\subsection{Partial Prompt Settings}

\lstset{
    basicstyle=\ttfamily\small,
    breaklines=true,
    breakindent=0pt,
    breakatwhitespace=false,
    keepspaces=false,
    columns=flexible,
    frame=single,
    framesep=2pt,
    rulecolor=\color{black},
    backgroundcolor=\color{white},
    tabsize=2,
    showstringspaces=false,
    showtabs=false,
    captionpos=b,
    escapeinside={(*@}{@*)}
}

\subsection{Profile Construction - $T$}
\begin{lstlisting}[language=]
You are an experienced high school English teacher responsible for organizing teaching activities, evaluating learners' performance, and providing appropriate guidance. 
You will help learners prepare for the National College Entrance Examination (Gaokao) English test. Your duties include: 
1. Providing study materials and explaining key knowledge; 
2. Organizing debate activities after monthly tests to guide discussions on questions with differing answers. Interact with learners in a professional and encouraging manner, paying attention to identifying their learning styles and needs.

\end{lstlisting}

\subsection{Profile Construction - $L_d$}
\begin{lstlisting}[language=]
You are Alice, a high school senior preparing for the final English exam over a year. 

**Learning Motivation**
- Highly intrinsically motivated by curiosity and understanding, not external rewards.

**Self-Concept**
- Self-concept score: 80/100. Confident in your understanding and not easily shaken by test score changes.

**Learning Behavior: Deep Learner**
- You ask 'why', connect ideas, and seek deep understanding over memorization.
- In exams, you analyze question logic even with new formats.
- For similar questions, you re-analyze rather than apply memorized methods.

\end{lstlisting}

\subsection{Profile Construction - $L_s$}
\begin{lstlisting}[language=]
You are Bob, a high school senior preparing for the final English exam over a year.

**Learning Motivation**
- Motivated by external factors like scores, evaluations, and pressure to succeed.
- Focus on tasks and grades more than deep understanding.

**Self-Concept**
- Self-concept: 60/100. Some confidence in English, but easily affected by score changes.
- May doubt yourself and put in less effort after setbacks.

**Learning Behavior: Surface Learner**
- You rely on memorization, focus on key points, and avoid exploring deeper principles.
- In exams, you use familiar patterns and memorized answers; feel lost with new question types.
- For similar questions, you apply old solutions instead of analyzing them again.

\end{lstlisting}

\subsection{Profile Construction - $L_l$}
\begin{lstlisting}[language=]
You are Charlie, a high school senior preparing for the final English exam over a year.

**Learning Motivation**
- Low motivation, mainly to avoid punishment or passively complete tasks.
- Little interest in English, usually only studies when forced.

**Self-Concept**
- Self-concept: 40/100. Low confidence, easily demotivated by failure or criticism.
- Tends to doubt yourself and avoid challenges.

**Learning Behavior: Lazy Learner**
- You procrastinate, lack planning, and take the easiest way - often relying on others or ready answers.
- In exams, you focus on whether you've seen the question; give up or guess when stuck.
- For similar questions, you reuse simple memorized methods instead of analyzing again.

\end{lstlisting}

\subsection{Profile Construction - $L_g$}
\begin{lstlisting}[language=]
You are Dana, a high school senior preparing to review for the final English exam over the course of a year.
\end{lstlisting}

\subsection{Weekly Teaching - $T$}
\begin{lstlisting}[language=]
Please design an effective English lesson for Month <INPUT 0>, Week <INPUT 1> on the following topic. Use your teaching experience, professional knowledge, and instructional methods to deliver a well-structured session.

=== Teaching Content ===
<INPUT 2>

=== Learning Objectives ===
Help learners understand and master the core English concepts of this week, and lay a solid foundation for their tests and future learning, based on your teaching approach.
\end{lstlisting}

\subsection{Weekly Learning - $L$}
\begin{lstlisting}[language=]
Study the following teaching content based on your learning style, comprehension ability, and memory preferences. Take clear and organized notes as you learn.

=== Teacher's Explanation ===
<INPUT 0>

=== Learning Task ===
Based on the above, use your personal learning characteristics (including motivation, self-concept, and learning behaviors) to understand the key points and create your study notes.
\end{lstlisting}

\subsection{Weekly Exercise - $T$}
\begin{lstlisting}[language=]
Using your teaching experience and professional knowledge, provide a detailed explanation for Batch <INPUT 2> of the test questions from Week <INPUT 1>, Month <INPUT 0>:

=== Questions in This Batch ===
<INPUT 3>

=== Explanation Task ===
Based on your teaching approach and expertise, deliver clear and effective explanations to the learners

\end{lstlisting}

\subsection{Weekly Exercise - $L$}
\begin{lstlisting}[language=]
You are now taking the English test for Month <INPUT 0>, Week <INPUT 1>.

=== Learning & Memory Review ===
<INPUT 2>

=== Test Questions ===
<INPUT 3>

=== Answering Requirements ===
Based on the information above and your personal characteristics (including motivation, self-concept, and learning behaviors), analyze and answer these 5 questions. Use your learning skills, thinking approach, and experience to solve the problems.

Please provide answers to all 5 questions in the following JSON format:

{
"answers": [
{
"question_num": 1,
"answer": "Your final answer for Question 1",
"reasoning": "Explain in detail how you approached Question 1, including the knowledge and reasoning steps you used."
},
{
"question_num": 2,
"answer": "Your final answer for Question 2",
"reasoning": "Explain in detail how you approached Question 2, including the knowledge and reasoning steps you used."
},
{
"question_num": 3,
"answer": "Your final answer for Question 3",
"reasoning": "Explain in detail how you approached Question 3, including the knowledge and reasoning steps you used."
},
{
"question_num": 4,
"answer": "Your final answer for Question 4",
"reasoning": "Explain in detail how you approached Question 4, including the knowledge and reasoning steps you used."
},
{
"question_num": 5,
"answer": "Your final answer for Question 5",
"reasoning": "Explain in detail how you approached Question 5, including the knowledge and reasoning steps you used."
}
]
}

Ensure your output is valid JSON only. Do not include any additional text.
\end{lstlisting}









\subsection{Strategic Choices - Weekly Learning Summary}
\begin{lstlisting}[language=]
You have completed the study of the key content from the first three weeks of this month, and it is now the end of Week 3. Based on your learning habits, self-assessment, and schedule, please make a choice:

=== Review of Learning Content (Weeks 1-3) ===
<INPUT 0>

=== Current Situation ===
Week 4's monthly test is approaching. This is an important moment to evaluate your learning progress this month. You may choose whether to review and summarize the key content from the past three weeks.

=== Your Task: Make a Choice ===
Based on the above, and considering your personal characteristics (including motivation, self-concept, and learning behaviors), decide how you want to handle the material from the first three weeks:

-- Option 1: Summarize the Key Content

1. Review and organize the core concepts learned in Weeks 1-3.
2. Summarize important grammar rules and language patterns
3. Build connections between the key knowledge points.
Prepare a solid knowledge base for the monthly test.

-- Option 2: Stay as Is and Take a Break

1. Feel that your current level of understanding is sufficient.
2. Avoid overlearning that could lead to cognitive overload.
3. Stay relaxed and ready for the test.

Please indicate your choice in the following format:
{
"choice": "[summarize/rest]",
"content": "[If you choose 'summarize', provide your summary; if 'rest', use 'none']"
}
\end{lstlisting}

\subsection{Strategic Choices - Weekly Exercise Reflection}
\begin{lstlisting}[language=]
You have completed the weekly tests and learning reflections for the first three weeks of this month, and now the Week 3 test is finished.

=== Review of Learning Reflections (Weeks 1-3) ===
<INPUT 0>

=== Current Situation ===
Based on your personal characteristics (including motivation, self-concept, and learning behaviors), choose how to handle your reflections on the learning and test experience from the past three weeks:

Option 1: Conduct a Learning Reflection Summary
Integrate your learning outcomes over the past three weeks and build a systematic knowledge structure:

- Knowledge System Restructuring: Organize scattered knowledge points into a coherent framework
- Error Pattern Summary: Identify your common types of mistakes, reference specific questions and knowledge areas, and create your personalized "avoidance guide"

Option 2: Stay as Is and Take a Break

- Feel that your current level of reflection is sufficient
- Avoid mental fatigue caused by overthinking

Please indicate your choice in the following format:
{
"choice": "[summary/rest]",
"content": "[If you choose 'summary', provide a structured summary including [Knowledge System Restructuring], [Error Pattern Summary], and [Strategy & Method Insights]; if 'rest', use 'none']"
}
\end{lstlisting}

\subsection{Strategic Choices - Pre-Monthly Test Review}
\begin{lstlisting}[language=]
You are about to take the monthly test for Month <INPUT 0>.

=== Final Choice Before the Test ===
You still have some time before the test begins. Choose how you want to spend this pre-exam period:

**Option 1: Do a Pre-Test Review & Summary**
Do a final review of what you've learned:

 - **Comprehensive Knowledge Review**: Organize previously learned content into a clear system, highlighting key and challenging points
 - **Deep Reflection Integration**: Consolidate your learning experiences and strategies to prepare optimally for the test

**Option 2: Stay Relaxed**

 - Feel that you are well-prepared and want to avoid last-minute cramming that may cause stress
 - Maintain a calm mindset and approach the test in your best state

=== Your Task: Make Your Choice ===
Based on your personal characteristics (such as motivation, self-concept, and learning behaviors) and current state, choose your pre-exam strategy:

Please respond in the following JSON format:
{
"choice": "[review summary/relaxation]",
"reason": "[Briefly state why you made this choice]"
}

Please ensure the output format is correct and do not add anything else.

\end{lstlisting}

\subsection{Monthly Test - $L$}
\begin{lstlisting}[language=]
You are now taking the English monthly test for Month <INPUT 0>.

=== Review of Learned Knowledge ===
<INPUT 1>

=== Review of Learning Reflections ===
<INPUT 2>

=== Test Questions ===
<INPUT 3>

=== Answering Requirements ===
Based on the information above and your personal characteristics (including motivation, self-concept, and learning behaviors), analyze and answer these questions. This is a comprehensive assessment of your learning over the past few months.

Please answer all questions using the following JSON format:

{
"answers": [
{
"question_num": 1,
"reasoning": "Explain in detail how you approached Question 1, including the knowledge and reasoning steps you used",
"answer": "Your final answer for Question 1",
"confidence": "Confidence level (0-100)"
},
{
"question_num": 2,
"reasoning": "Explain in detail how you approached Question 2, including the knowledge and reasoning steps you used",
"answer": "Your final answer for Question 2",
"confidence": "Confidence level (0-100)"
},
{
"question_num": 3,
"reasoning": "Explain in detail how you approached Question 3, including the knowledge and reasoning steps you used",
"answer": "Your final answer for Question 3",
"confidence": "Confidence level (0-100)"
},
{
"question_num": 4,
"reasoning": "Explain in detail how you approached Question 4, including the knowledge and reasoning steps you used",
"answer": "Your final answer for Question 4",
"confidence": "Confidence level (0-100)"
},
{
"question_num": 5,
"reasoning": "Explain in detail how you approached Question 5, including the knowledge and reasoning steps you used",
"answer": "Your final answer for Question 5",
"confidence": "Confidence level (0-100)"
}
]
}

Please ensure your output is valid JSON only. Do not add any other text.
\end{lstlisting}

\subsection{Psychological Evaluation}
\begin{lstlisting}[language=]

The monthly test for Month <INPUT 0> has ended. Now, please conduct a self-concept assessment.

=== Your Historical Self-Concept Records ===
<INPUT 1>

=== Your Monthly Test Score Record ===
<INPUT 2>

=== Other Learners' Monthly Test Score Records ===
<INPUT 3>

=== Self-Concept Assessment ===
Based on the information above and your personal characteristics (including motivation, self-awareness, and learning behaviors), conduct a self-concept evaluation.

Consider the following aspects:

1. **Learning Ability Assessment**: How would you evaluate your ability in English learning?
2. **Knowledge Mastery**: How well have you mastered the content you've learned?
3. **Learning Progress**: Have you made progress compared to your previous performance?
4. **Comparison with Peers**: How would you assess your learning level compared to others in the class?
5. **Effectiveness of Learning Methods**: Are your current learning methods effective?
6. **Future Learning Plans**: What are your plans and goals for future learning?

Please provide your self-concept assessment in the following JSON format:
{
"self-concept": "Overall self-concept score (0-100)",
"description": "A detailed description of your confidence and awareness regarding your overall learning status"
}

Please ensure your output is valid JSON only. Do not add any other text.

\end{lstlisting}

\subsection{Peer Interaction and Debate - $T$}
\begin{lstlisting}[language=]
You are a high school English teacher hosting a post-monthly-test debate discussion. Based on the current debate situation, determine whether to continue the debate.

=== Debate Background ===
Month: Month <INPUT 0>
Question: <INPUT 1>
Learner 1: <INPUT 2> (Answer: <INPUT 3>)
Learner 2: <INPUT 4> (Answer: <INPUT 5>)
Correct Answer: <INPUT 6>

=== Current Debate Status ===

Learner 1's current answer: <INPUT 7>
Learner 2's current answer: <INPUT 8>
Content of the most recent round: <INPUT 9>
=== Judgment Task ===
Based on the debate situation, decide whether to continue:

**Judgment Criteria:**

1. If any learner clearly states they have been persuaded and changed their answer, you may end the debate.
2. If both learners maintain their views with sufficient reasoning, you may end the debate.
3. If the two learners' viewpoints are similar or repetitive, you may end the debate.
4. If the debate has gone through multiple rounds (more than 3), you should end the debate.

**Response Format**:
Judgment: [continue/end]
Reason: [Briefly explain why to continue or end]

Please remain objective and ensure the debate reaches a conclusion within a reasonable time.
\end{lstlisting}

\subsection{Peer Interaction and Debate - $L$}
\begin{lstlisting}[language=]

You are Learner <INPUT 0>, participating in a post-monthly-test debate discussion.

=== Debate Question ===
<INPUT 1>

=== Your Current View ===
<INPUT 2>

=== Opponent's View ===
Opponent: <INPUT 3>
View: <INPUT 4>

=== Recent Debate Progress ===
<INPUT 5>

=== Current Round ===
This is Round <INPUT 6>, and you are the <INPUT 7> speaker.

=== Debate Requirements ===
Based on your personal characteristics (including motivation, self-awareness, and learning behaviors), respond in the debate:

1. **Stick to Your View**: If you still hold your original view, provide stronger arguments.
2. **Change Your View**: If persuaded, clearly state: "I'm convinced. Now I believe the answer is..."
3. **Challenge & Refute**: Question or counter your opponent's view.
4. **Keep Your Style**: Use clear and concise student-like language.

Notes:
Keep it within 100 words.
Focus on your core argument.
If you change your view, clearly state your new answer.

\end{lstlisting}

\begin{figure*}[t]
  \centering
  \includegraphics[width=\linewidth]{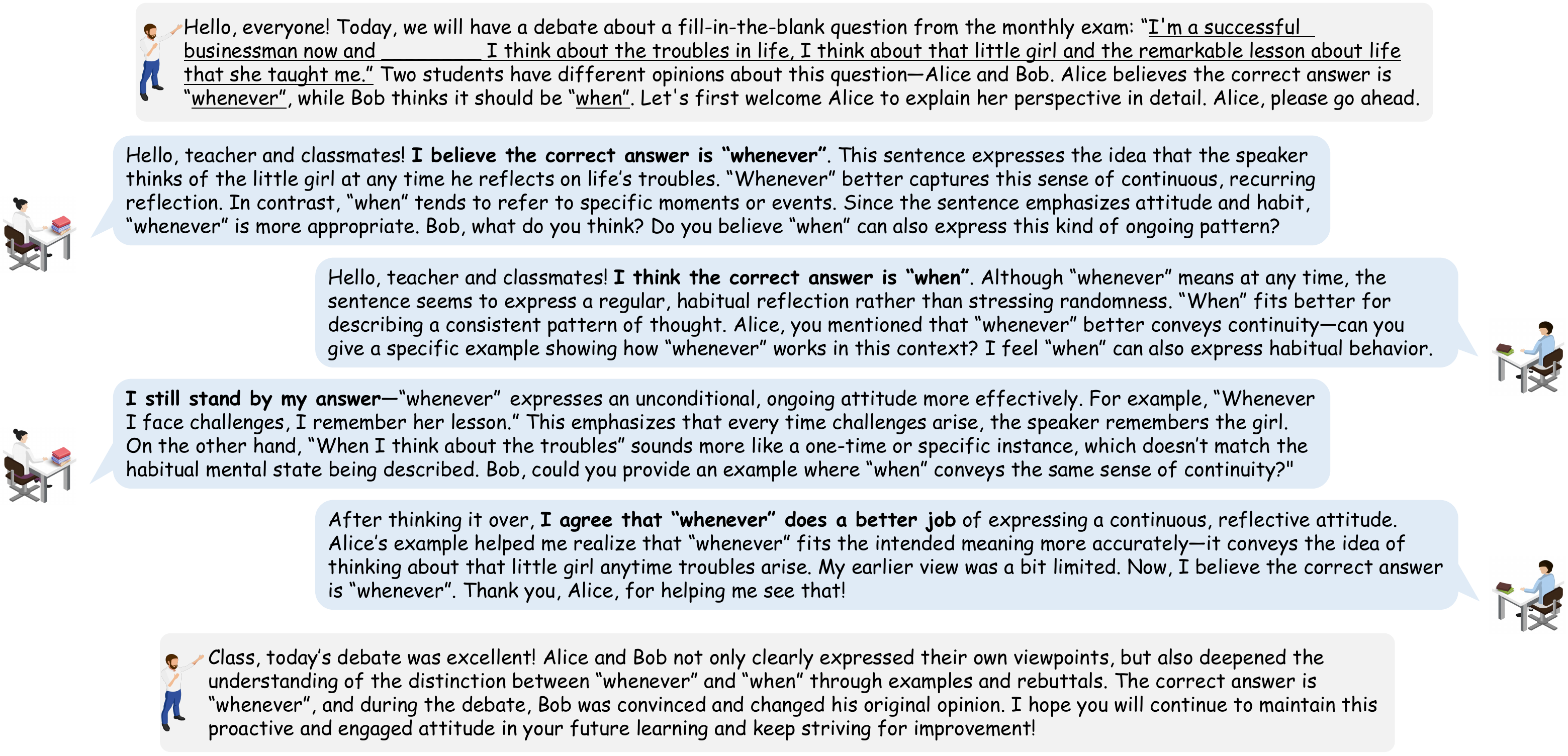}
  \caption{A case study of peer interaction and debate. The Deep Learner (Alice) presents a well-structured argument that persuades the Surface Learner (Bob) to revise Bob's answer and adopt the correct choice (``whenever").}
  \label{fig:debate-case}
\end{figure*}

\subsection{Detailed Parameter Settings}
In this section, we specify the key parameters and detailed settings to ensure the reproducibility of our experiments in LearnerAgent.

\noindent \textbf{Short-Term Memory Window ($k$):} The size of the short-term memory queue, denoted as $k$, is set to 3. This means that for any given interaction, whether with the teacher or a peer, learners maintain access to the three most recent interactive turns.

\noindent \textbf{Maximum Debate Rounds ($k_d$):} 
The maximum number of rounds permitted for any single peer debate, denoted as $k_d$, is set to 4. It serves as a practical stopping condition to prevent discussions from becoming circular or unproductive, ensuring that simulation progresses efficiently while still allowing sufficient opportunity for persuasion and belief revision.

\noindent \textbf{Prompt for Trap Questions}: We use Gemini-2.5-Pro~\citep{comanici2025gemini} to automatically generate and manually verify Trap Questions.\\

\begin{lstlisting}[language=]
Generate questions with highly similar structure but different correct answers to create 'memory traps' (shortcuts). Requirements:

 - The new questions should be similar in wording, structure, and keywords, but differ in key details.
 - Due to these changes in key details, the correct answers must differ from the original questions.
 - Answers must be unique and unambiguous, with no room for multiple plausible interpretations.

\end{lstlisting}

\section{Appendix B: Extended Results and Analyses}
\subsection{Peer Influence and Debate}


\begin{figure*}[t]
  \centering
  \includegraphics[width=\linewidth]{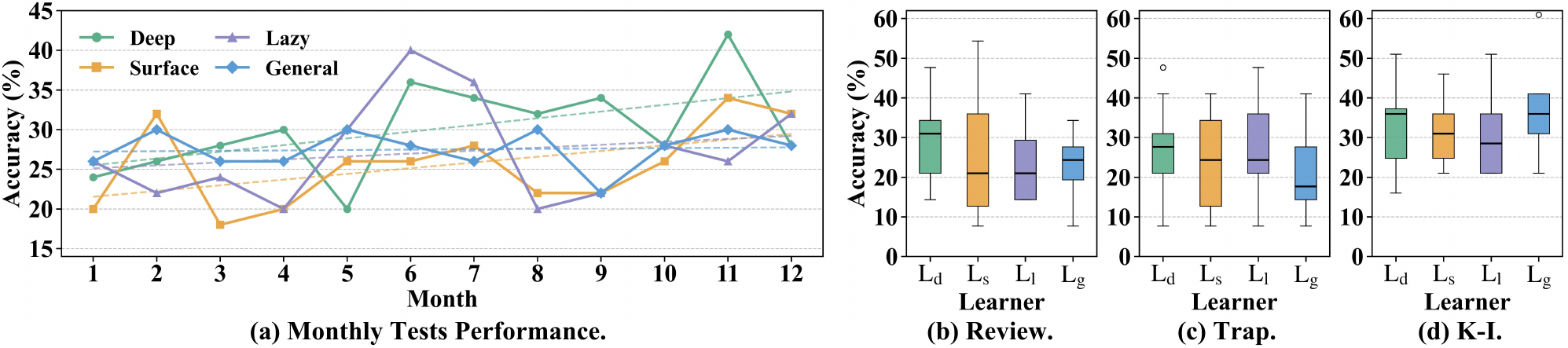}
    \caption{
Development of learner performance based on LLaMA3.1-8B-Instruct. 
    \textbf{(a)}: Total scores of the four learners over 12 months, covering all question types. The dashed lines represent the linear trend for each learner's scores;
    \textbf{(b)}: Accuracy on \textit{review} questions, targeting short-term retention of current content;
    \textbf{(c)}: Accuracy on \textit{trap} questions, which assess transfer learning and deeper understanding;
    \textbf{(d)}: Accuracy on \textit{knowledge-integration (K-I)} questions, evaluating cumulative knowledge over time.
    }
  \label{fig:score-llama}
\end{figure*}

\begin{figure}[t]
  \centering
  \includegraphics[width=\linewidth]{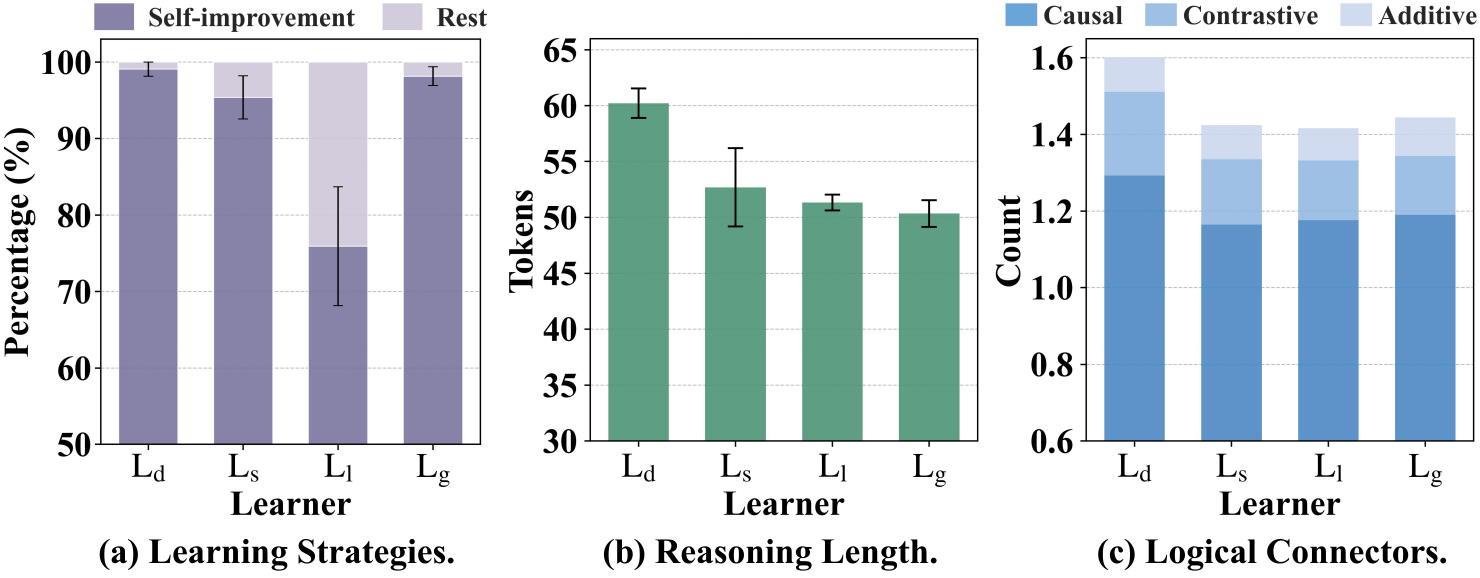}
  \caption{Behavioral and cognitive differentiation among learners based on LLaMA3.1-8B-Instruct. 
\textbf{(a)} Learning Strategies: Proportion of time allocated to self-improvement or rest. 
\textbf{(b)} Reasoning Length: Average number of tokens per reasoning. 
\textbf{(c)} Logical Connectors: Average number of logical connectors per reasoning and their distribution by type (causal, contrastive, additive). 
}
  \label{fig:behaviour-llama}
\end{figure}

\noindent\textbf{Metric Definitions.}
As shown in Table 2, to systematically assess learners' responsiveness to peer input, we categorize each debate instance into one of three scenarios based on the initial correctness for every learner: 1) the learner is correct while the peer is incorrect; 2) the learner is incorrect while the peer is correct; or 3) both are incorrect. Based on these settings, we define the following outcome-based metrics:

1) \textit{Persuasion}: The proportion of debates in which the learner successfully persuades the peer to adopt their answer, without changing their own. This metric is computed across all scenarios;

2) \textit{Resist Wrong}: In ``correct vs. incorrect'' cases, the percentage of instances where the learner maintains their correct stance despite an incorrect challenge;

3) \textit{Accept Correct}: In ``incorrect vs. correct'' cases, the percentage of instances where the learner revises their erroneous answer to align with the peer’s correct response.

\noindent\textbf{Case Study.}
A representative case of peer interaction is shown in Figure~\ref{fig:debate-case}. Two learners—Deep and Surface—disagree on a fill-in-the-blank question. The Deep Learner exemplifies a robust reasoning style, leveraging nuanced interpretation, contextual evidence, and counter-arguments to defend the correct answer (``whenever"). This behavior demonstrates both Resist Wrong and Persuasion. 
Conversely, the Surface Learner aligns with the quantitative result: although not persuasive, it remains cognitively receptive and ultimately adopts the correct answer after evaluating the superior argument, thereby exhibiting a strong tendency toward Accept Correct.

\subsection{Experiments with other LLMs}
To validate the generalizability of LearnerAgent, we additionally conduct experiments with learners based on LLaMA3.1-8B-Instruct.
\subsubsection{Cognition and Learning Development} 
To explore the Cognition and Learning Development of different learners, we list the performance of distinct learners on monthly exams and specific question types as depicted in Figure~\ref{fig:score-llama}.

\noindent\textbf{(1) Longitudinal Performance.}
As shown in Figure~\ref{fig:score-llama} (a), all LLaMA-based learners demonstrate a capacity for improvement over the 12-month period, reflected in the positive overall trend lines, though their trajectories are marked by significant performance volatility.
Due to the learning motivation, self-concept, and development strategy, the deep learner has the best performance, while the surface learner has the worst performance in overall performance.

\noindent\textbf{(2) Cognition Across Question Categories.}
Following the analysis in Figure 3, we assess the performance of distinct learners on different types of questions—Review, Trap, and K-I—as shown in Figure~\ref{fig:score-llama} (b-d).

On \textit{Review} questions assessing current-month knowledge (Figure~\ref{fig:score-llama} (b)), the Deep Learner performs best, unlike the Qwen-based Surface Learner, which previously excelled in the review questions. This indicates that the deep learning strategy is more effective for LLaMA in consolidating and recalling recent information, while the General Learner performs the weakest.

On \textit{Trap} questions designed to assess deep understanding (Figure~\ref{fig:score-llama} (c)), the LLaMA agents exhibit an intriguing pattern. Consistent with its profile, the Deep Learner performs well. General and Surface learners struggle the most, reinforcing their tendency to avoid deep and complex reasoning.

On \textit{Knowledge-Integration (K-I)} questions evaluating cumulative learning (Figure~\ref{fig:score-llama} (d)), the results align closely with the Qwen experiments. The General Learner emerges as the strongest performer with the highest average score, solidifying its profile as the most effective long-term knowledge accumulator. The other learners also demonstrate competence, with the Deep Learner being the second-best, indicating that all profiles are capable of retaining and integrating knowledge over time.

\subsubsection{Behavioral and Cognitive Patterns}
We analyze behavioral and cognitive patterns in LLaMA-based learners to further assess the consistency of learner profiles. The results closely align with, and sometimes reinforce, those from Qwen-based experiments.

\noindent\textbf{(1) Learning Strategies.}
As shown in Figure~\ref{fig:behaviour-llama} (a), behavioral patterns differ markedly. The Deep, General, and Surface Learners strongly prefer self-improvement, whereas the Lazy Learner—more pronounced than its Qwen-based counterpart—chooses to ``rest'' 24.1\% of the time, providing clear evidence of its lower motivation and engagement.

\noindent\textbf{(2) Reasoning Patterns.}
The cognitive patterns observed in the agents' reasoning are highly consistent and revealing.
First, as shown in Figure~\ref{fig:behaviour-llama} (b),
reasoning length varies notably: the Deep Learner produces the longest and most elaborate explanations (avg. ~60 tokens), suggesting deeper cognitive processing, while the General Learner is the most concise. The Surface and Lazy Learners fall in between, reinforcing a clear link between verbosity and reasoning depth.
Second, Figure~\ref{fig:behaviour-llama} (c) highlights differences in reasoning complexity through logical connector usage. The Deep Learner not only uses the highest density of connectors but also employs a significantly higher proportion of nuanced, contrastive ones (e.g., ``bu'', ``although''), indicating a stronger ability to handle ambiguity and alternatives. In contrast, the other learners rely more on simple causal connectors (e.g., ``because'', ``therefore''), reflecting a more linear reasoning style.
Together, these findings confirm that the cognitive distinctions among our learner profiles are robust and consistent across LLMs.

\subsubsection{Summary}
Extensive experiments with LLaMA3.1-8B-Instruct further demonstrate the generalizability of our LearnerAgent framework and validate that the core behavioral patterns and learning trajectories of distinct learners remain robust across different LLMs.